\documentclass[a4paper,10pt]{article}

\usepackage{latexsym}
\usepackage{mathrsfs}
\usepackage{amsfonts,amsmath,amssymb}
\usepackage{indentfirst}
\usepackage{subeqnarray}
\usepackage{verbatim}
\usepackage[pdftex]{graphicx}
\usepackage{epstopdf}
\usepackage{stmaryrd}
\usepackage{multirow}
\usepackage{diagbox}
\usepackage{subcaption}
\usepackage{bm}

\newcommand{\bx}{\mbox{\boldmath{$x$}}}

\newcommand{\by}{\mbox{\boldmath{$y$}}}

\newtheorem{theorem}{Theorem}[section]

\newtheorem{example}[theorem]{Example}

\numberwithin{equation}{section}

\usepackage[pdftex,unicode,colorlinks,linkcolor=blue]{hyperref}

\begin{document}

\begin{center}
\Large\bf DeepONet Augmented by Randomized Neural Networks\\ for Efficient Operator Learning in PDEs
\end{center}

\begin{center}
Zhaoxi Jiang\footnote{School of Mathematics and Statistics, Xi'an Jiaotong University, Xi'an, Shaanxi 710049, P.R. China.   E-mail: {\tt zhaoxi9614@stu.xjtu.edu.cn}.},\quad 
Fei Wang\footnote{School of Mathematics and Statistics, Xi’an Jiaotong University, Xi’an, Shaanxi 710049, China. The work of this author was partially supported by the National Natural Science Foundation of China (Grant No.\ 92470115), Tianyuan Fund for Mathematics of the National Natural Science Foundation of China (Grant No. 12426105).  Email: {\tt feiwang.xjtu@xjtu.edu.cn}.}
\end{center}

\medskip
\begin{quote}
  {\bf Abstract.} Deep operator networks (DeepONets) represent a powerful class of data-driven methods for operator learning, demonstrating strong approximation capabilities for a wide range of linear and nonlinear operators. They have shown promising performance in learning operators that govern partial differential equations (PDEs), including diffusion-reaction systems and Burgers’ equations. However, the accuracy of DeepONets is often constrained by computational limitations and optimization challenges inherent in training deep neural networks. Furthermore, the computational cost associated with training these networks is typically very high. To address these challenges, we leverage randomized neural networks (RaNNs), in which the parameters of the hidden layers remain fixed following random initialization. RaNNs compute the output layer parameters using the least-squares method, significantly reducing training time and mitigating optimization errors.
  In this work, we integrate DeepONets with RaNNs to propose RaNN-DeepONets, a hybrid architecture designed to balance accuracy and efficiency. Furthermore, to mitigate the need for extensive data preparation, we introduce the concept of physics-informed RaNN-DeepONets. Instead of relying on data generated through other time-consuming numerical methods, we incorporate PDE information directly into the training process.
  We evaluate the proposed model on three benchmark PDE problems: diffusion-reaction dynamics, Burgers' equation, and the Darcy flow problem. 
  Through these tests, we assess its ability to learn nonlinear operators with varying input types. When compared to the standard DeepONet framework, RaNN-DeepONets achieves comparable accuracy while reducing computational costs by orders of magnitude. These results highlight the potential of RaNN-DeepONets as an efficient alternative for operator learning in PDE-based systems.

\end{quote}


\medskip

\section{Introduction}

Partial differential equations (PDEs) are fundamental in many fields, such as physics, economics, and image processing. Numerous methods have been proposed to obtain their numerical solutions. Traditional methods, such as finite element, finite volume, and spectral methods, often achieve satisfactory accuracy for these problems. However, these methods tend to be case-specific, meaning they need to be recomputed when facing a new equation, even if the differences between the old and new equations are minimal. To address this, operator learning methods have been developed to learn mappings between function spaces. While these operators act on infinite-dimensional functions in principle, practical implementations leverage discretization-invariant parameterizations to handle finite-dimensional approximations. These operators, often learned implicitly from data, act as unified solution mappings that can solve a class of PDEs with the same operator, eliminating the need for equation-specific modifications.

Artificial neural networks (ANNs) have demonstrated to be highly effective across a wide range of problems, including image segmentation, time series forecasting, and natural language processing (\cite{segment1, segment2, series1, series2, NLP}), owing to their strong approximation capabilities (\cite{Un}). The universal operator approximation theorem, developed in \cite{UnTheroy}, demonstrates that a neural network with a single hidden layer can accurately approximate any nonlinear continuous functional (\cite{funapp1, funapp2}) and operator (\cite{operapp}). As a result, recent research has increasingly focused on using neural networks to learn the operators of nonlinear PDEs. The neural operator, introduced in \cite{Noper}, approximates operators through an iterative architecture, employing neural networks to update the operator at each iteration. Three prominent frameworks for learning PDE operators include the Fourier Neural Operator (FNO) (\cite{FNO1}), Deep Operator Networks (DeepONets) (\cite{DeepONet1}), and Physics-Informed Neural Operators (PINO) (\cite{PINO1}). FNO is a novel deep learning architecture that follows the neural operator approach but utilizes Fourier integral operators at each iteration. PINO enhances the generalization capability of the model by embedding physics constraints through PDE residuals and calculating model derivatives, thus reducing data dependency.

DeepONets, introduced by Lu et al. (\cite{DeepONet1}), are inspired by the generalized universal approximation theorem for operators (\cite{UnTheroy}). The architecture of this approach comprises two primary sub-networks: the branch network and the trunk network. DeepONets have demonstrated strong performance across a variety of applications (\cite{deepapp1, deepapp2, deepapp3}). In comparison to the FNO, DeepONets offer greater flexibility in the structure of the branch network, facilitating a wide range of extensions. Physics-informed DeepONets (\cite{Pideep}) enhance the generalization ability of the basic DeepONets and reduce data dependency by incorporating physics information into the neural network. POD-DeepONets (\cite{PODdeep}), which leverage Proper Orthogonal Decomposition (POD) bases for the trunk network, improve model accuracy while significantly reducing the number of parameters. Multi-task DeepONets (\cite{multideep}) have demonstrated their ability to simultaneously handle multiple types of inputs. However, methods such as PINO, FNO, and DeepONets still face challenges, as they involve solving nonlinear and non-convex optimization problems with a large number of parameters during training. Existing algorithms, such as Adam (\cite{Adam}) and regularization techniques (\cite{regular}), often lead to high computational costs and significant optimization errors, limiting the accuracy and broader applicability of these approaches.

In addition to these neural operator-based methods, several alternative techniques have been developed to learn PDE operators, showing potential in addressing the aforementioned challenges. The reduced basis method (\cite{redu1, redu2}), a classical approach, reduces the dimensionality of the problem space by constructing an empirical basis, thereby simplifying the problem and enabling a mapping from the input function space to the output function space of PDEs. In the encoder-decoder paradigm, Oliva et al. (\cite{redu3}) developed a function-to-function regression (FFR) method, where the input and output spaces are represented by truncated orthogonal bases. The random feature method (RFM) (\cite{RFM1, RFM2}) is another approach that serves as a mapping between finite-dimensional spaces and is particularly well-suited for learning PDE operators. Several studies have explored the application of RFM to operator learning, with promising results (\cite{RFMOL1, RFMOL2}).

Randomized Neural Networks (RaNNs), proposed in \cite{RNN1, RNN2, RNN3}, have emerged as an effective strategy for reducing computational time and minimizing optimization errors. Based on the architecture of Fully Connected Neural Networks (FCNs), RaNNs differ in the way their parameters are handled. Unlike the vanilla model, the parameters of the links between the hidden layers in RaNNs are randomly sampled from certain distributions and fixed during the training process, while the parameters for the connections between the last hidden layer and the output layer are solved using a least-squares method. This approach allows RaNNs to solve only a least-squares problem during training, significantly reducing computational resource requirement and optimizing error. RaNNs have shown excellent performance in solving quantitative PDEs (\cite{RNNchen, RNNdong, RNNapp1, RNNapp2, RNNapp3, RNNapp4}) with reduced time cost and high accuracy. 

In this work, we propose a novel integration of RaNNs into the DeepONet framework, introducing RaNN-DeepONets. Taking advantage of the inherent flexibility of DeepONets, we replace the branch network with RaNNs and use random basis functions for the trunk network. The resulting model exhibits strong approximation capabilities for a wide range of highly nonlinear PDE operators. Its accuracy is comparable to that of the DeepONet model for many problems and even surpasses it in certain cases. Moreover, compared to DeepONets— which often require hours of training and large numbers of parameters— the training process of RaNN-DeepONets takes just a few minutes.

The remainder of this paper is structured as follows. In Section \ref{deeponet}, we introduce the DeepONet framework and some of its extensions that will be utilized in the subsequent sections. In Section \ref{RNN}, we discuss the architecture of RaNNs and the initialization strategy for bias parameters. In Section \ref{RaNN-DeepONet}, we integrate DeepONets with RaNNs to propose RaNN-DeepONets and reformulate the training process as solving a linear system. The concept of physics-informed RaNN-DeepONets is also introduced. Section \ref{sec:numericalexample} provides numerical examples that validate the approximation capabilities of RaNN-DeepONets for various nonlinear operators. Finally, we conclude the paper with a summary of the proposed method.

\section{Deep Operator Networks}
\label{deeponet}

DeepONets are a class of neural operator approaches designed to learn an operator $\mathcal{G}$ that maps from an input $f$ (such as a source term, initial condition, or domain) to a target function $u$. A trained DeepONet can be represented as $\mathcal{G}(f)(\bm{y})$, where $\by$ denotes the coordinate of an arbitrary point within the domain. The output of the network is a function that approximates the target function $u(\by)$. While DeepONets can take various forms, this work primarily focuses on vanilla DeepONets, physics-informed DeepONets, and several simple extensions aimed at improving accuracy.

\subsection{Vanilla DeepONets}\label{vanDeep}

The vanilla DeepONet architecture includes two variants: stacked and unstacked. The unstacked version uses a single neural network as branch net which combined with a multi-layer trunk net, as proposed by Lu et al. (\cite{DeepONet1}), building upon the generalized universal approximation theorem for operators originally established by Chen \& Chen (\cite{UnTheroy}). 

To apply vanilla DeepONets for solving PDEs, the input function $f$ must first be discretized by sampling its values at predefined sensor locations, which are then fed into the branch network. In contrast, the trunk network operates on continuous coordinate inputs.
By selecting a set of sensors \(\{\bx_1, \bx_2, \dots, \bx_m\}\), $f$ is mapped to its pointwise evaluations on these sensors, which are represented as \(\{f(\bx_1), f(\bx_2), \dots, f(\bx_m)\}\). This finite-dimensional discretized representation $\bm{f}$ serves as the actual input to the branch network.
Assuming there are $N$ realizations of $\bm{f}$, and for each realization, $q$ collocation points are selected to evaluate $u$, the data for the DeepONet represented as $\{\bm{f}^{(n)},\{\bm{y}^{(n)}_j,u^{(n)}_j\}^q_{j=1}\}^N_{n=1}$, where $\bm{f}^{(n)}=[f^{(n)}(\bx_1),f^{(n)}(\bx_2),\cdots,f^{(n)}(\bx_m)]$.

\begin{figure}[h]
	\centering
	\includegraphics[width=0.8\textwidth]{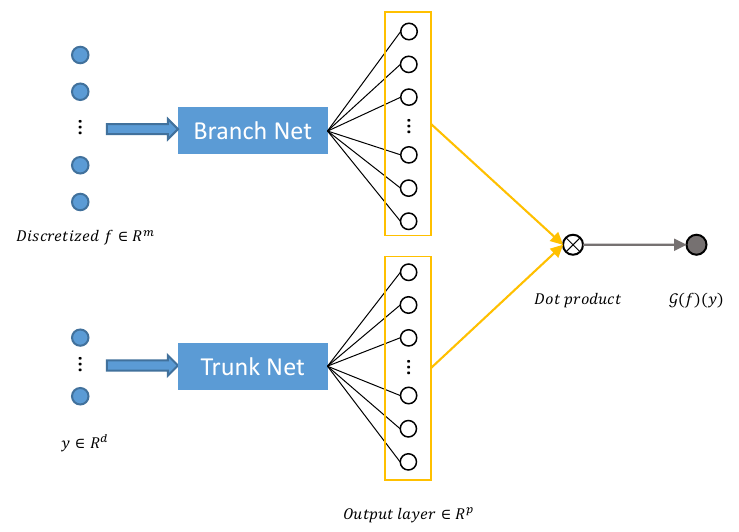}
	\caption{Architecture of vanilla DeepONets}
	\label{fig_deep}
\end{figure}

As shown in Figure \ref{fig_deep}, vanilla DeepONets consist of two sub-networks: the branch network and the trunk network. The input to the branch network is the $m$-dimensional discretized vector $\bm{f}$, which is mapped into $\bm{b}(\bm{f})$ as its output. The input to the trunk net is $\bm{y}$, which represents the coordinates of the points of interest, and the output is $\bm{t}(\bm{y})$. Since $\bm{b}(\bm{f})$ and $\bm{t}(\bm{y})$ have the same dimensionality $p$, a dot product can be computed to evaluate $\mathcal{G}(\bm{f})(\bm{y})$. Thus, the output of the vanilla DeepONets is expressed as
\begin{equation}
	\mathcal{G}(\bm{f})(\bm{y})=\sum_{i=1}^{p} b_i(\bm{f}) t_i(\bm{y}). 
\end{equation}
Since $\by$ typically represents the independent variables in a PDE, it is often a low-dimensional vector, which is why simple fully connected neural networks are commonly used as the trunk network for DeepONets. In contrast, the branch network is more flexible, and various neural network models such as FCNs, convolutional neural networks and graph neural networks (\cite{GNN1}) can be used, depending on the properties of $\bm{f}$. Let $\mathcal{G}_\theta(\bm{f})(\bm{y})$ represent a DeepONet parameterized by $\theta$, and the corresponding loss function $\mathcal{L}(\theta)$ is:
\begin{equation}\label{MSEloss}
	\mathcal{L}(\theta)=\frac{1}{Nq}\sum_{n=1}^{N}\sum_{j=1}^{q}[\mathcal{G}_\theta(\bm{f}^{(n)})(\bm{y}^{(n)}_j)-u^{(n)}(\bm{y}^{(n)}_j)]^2.
\end{equation}

\subsection{Physics-informed DeepONets}\label{PIdeep}

\begin{figure}[h]
	\centering
	\includegraphics[width=0.8\textwidth]{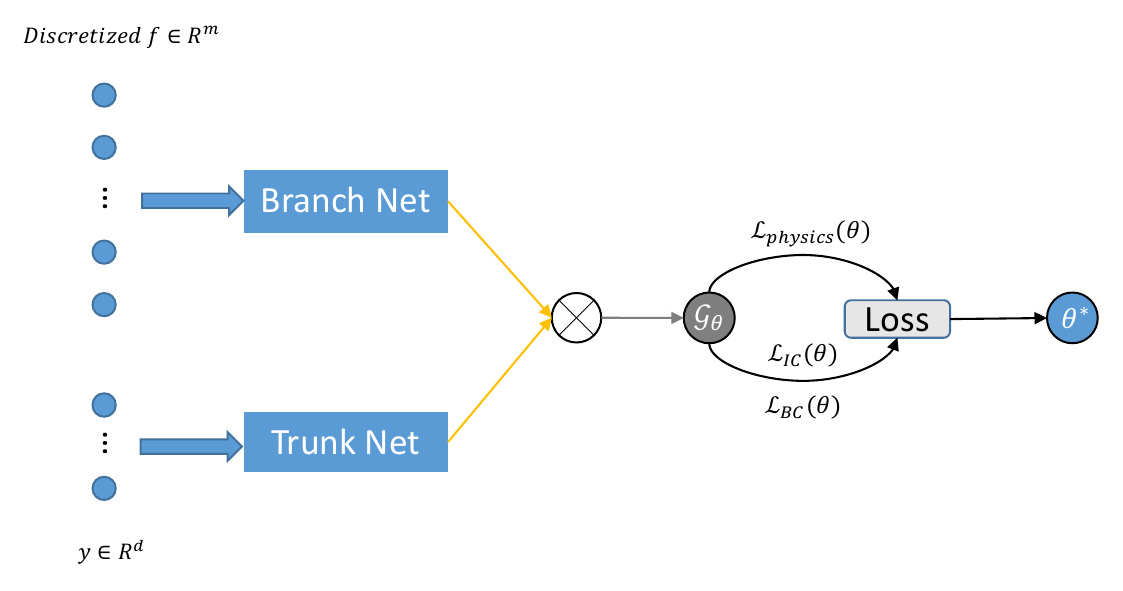}
	\caption{Architecture of physics-informed DeepONets}
	\label{fig_pideep}
\end{figure}

Proposed by Wang et al. (\cite{Pideep}), physics-informed DeepONets follow the general architecture of vanilla DeepONets with modifications to the loss function. As shown in Figure \ref{fig_pideep}, physics-informed DeepONets divide the loss into two components: boundary/initial loss and physics loss, denoted by $\mathcal{L}_{BC/IC}(\theta)$ and $\mathcal{L}_{physics}(\theta)$, respectively.

Consider the following PDE problem:
\begin{equation}\label{PIPDE}
	  \begin{aligned}
		&\mathcal{N}(u)=f, \ \bx \in \Omega, \\
		&\mathcal{B}(u)=u, \ \bx \in \partial\Omega,
		\end{aligned}
\end{equation}
where $\mathcal{N}$ and $\mathcal{B}$ are operators mapping $u$ to source term and boundary/initial condition, respectively. During the training process, the physics-informed DeepONets minimize the loss function
\begin{equation}
	\mathcal{L}(\theta)=\mathcal{L}_{physics}(\theta)+\lambda \mathcal{L}_{BC/IC}(\theta),
\end{equation}
where $\lambda$ is a hyperparameter that controls the weight of boundary loss. The boundary/initial condition loss $\mathcal{L}_{BC/IC}(\theta)$ is the same as the loss in vanilla DeepONets and the physics loss is given by
\begin{equation}\label{PIloss}
	\mathcal{L}_{physics}(\theta)=\frac{1}{Nq}\sum_{n=1}^{N}\sum_{j=1}^{q}[\mathcal{N}(\mathcal{G}_\theta(\bm{f}^{(n)})(\bm{y}^{(n)}_j))-f^{(n)}(\bm{y}^{(n)}_j)]^2.
\end{equation}

By incorporating physics information into the model, physics-informed DeepONets reduce the need for solution values within the domain and enhance generalization capabilities. However, this approach comes at the cost of higher time complexity compared to vanilla DeepONets, as it requires the use of automatic differentiation to compute derivative information, which can be computationally expensive during training.

\subsection{Hard-constraint Boundary Conditions}\label{HBC}

Several methods have been proposed to enhance the accuracy of DeepONets, including the hard-constraint boundary conditions (BCs) method introduced by Lu et al. (\cite{PODdeep}). This method enforces Dirichlet or periodic boundary conditions as hard constraints within DeepONets.

Enforcing Dirichlet BCs in neural networks has been widely utilized in physics-informed neural networks (\cite{PINN, PINN2}). Consider a PDE defined on the domain $\Omega$ with boundary conditions given by $g$. To ensure that the DeepONet output satisfies these BCs automatically, a surrogate solution is introduced
\begin{equation}\label{DBCs}
\mathcal{G}(\bm{f})(\bm{y})=c(\bm{y})\widetilde{\mathcal{G}}(\bm{f})(\bm{y})+g(\bm{y}),
\end{equation}
where $\widetilde{\mathcal{G}}(\bm{f})(\bm{y})$ is the output of the DeepONet and $c(\bm{y})$ satisfies the following conditions:
\begin{equation}
	\left\{\begin{aligned}
		&c(\bm{y})=0, \ \bm{y}\in \partial\Omega, \\
		&c(\bm{y})>0, \ otherwise. 
	\end{aligned}\right.
\end{equation}

For enforcing periodic BCs as hard constraints (\cite{PINN2, pbc}), Fourier basis functions are employed for feature expansion. Consider a 1-dimensional ordinary differential equation (ODE)
\begin{equation}
	\mathcal{N}(u(x))=f(x), \ x\in (0,T)
\end{equation}
with periodic boundary conditions:
\begin{equation}
	\begin{aligned}
		u(0)&=u(T), \\
		u_x(0)&=u_x(T), 
	\end{aligned}
\end{equation}
where $\mathcal{N}$ is a differential operator mapping $u(x)$ into $f(x)$. Periodic BCs can be enforced as hard constraints by expanding the features on $x$. For example,
\begin{equation}
	x\rightarrow \{\cos(\omega x),\sin(\omega x),\cos(2\omega x),\sin(2\omega x)\},
\end{equation}
where $\omega=\frac{2\pi}{T}$. Since the input of trunk net is periodic, the output of DeepONet will also be periodic automatically. A similar approach for enforcing periodic BCs in higher dimensions can be found in \cite{PODdeep}, which extends the 1-dimensional method to higher-dimensional cases.

\section{Randomized Neural Networks}
\label{RNN}

In this section, we introduce the framework of Randomized Neural Networks (RaNNs), which are primarily based on FCNs in this work. We also present the bias initialization strategy for RaNNs.

\subsection{Architecture of Randomized Neural Network}
\label{ArRNN}

Consider a feedforward neural network (FNN) with a single hidden layer, structured as follows: an input layer of $m$ neurons, a hidden layer with 
$k$ neurons, and an output layer of $p$ neurons. This architecture can be mathematically expressed as
\begin{equation}
	\bm{u}=\bm{W}^{(2)}\sigma(\bm{W}^{(1)}\bm{x}+\bm{b}^{(1)})+{\bm{b}^{(2)}},\label{FNN}
\end{equation} 
where $\bm{x}\in \mathbb{R}^m$ and $\bm{u}\in \mathbb{R}^p$ are the input and output of FNN. $\bm{W}^{(1)}\in \mathbb{R}^{k\times m}$ and $\bm{W}^{(2)}\in \mathbb{R}^{p\times k}$ are the weight matrices for the input-to-hidden and hidden-to-output layers, respectively. $\bm{b}^{(1)}\in \mathbb{R}^{k}$ and $\bm{b}^{(2)}\in \mathbb{R}^{p}$ are the biases of the hidden and output layers. The function $\sigma(\cdot)$ is the activation function.

\begin{figure}[h]
	\centering
	\includegraphics[width=0.8\textwidth]{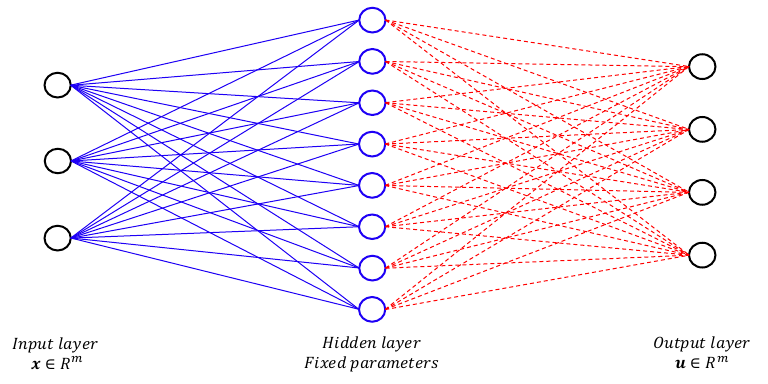}
	\caption{Architecture of Randomized Neural Network}
	\label{fig_RNN}
\end{figure}

RaNNs follow the architecture of FCNs, with key differences in the handling of parameters. As shown in Figure \ref{fig_RNN}, the parameters associated with the blue units are randomly selected and fixed during the training process. For simplicity, assume that the bias of the output layer $\bm{b}^{(2)}$ is zero. The remaining parameter is $\bm{W}^{(2)}$, indicated by the red dotted lines. Let $\bm{h}$ be the output of the hidden layer,  then the RaNN model can be rewritten as:
\begin{equation}
	\begin{aligned}
		&{u}=\bm{W}^{(2)}\bm{h},\\	
		&\bm{h}=\sigma(\bm{W}^{(1)}\bm{x}+\bm{b}^{(1)}).\label{RaNN}
	\end{aligned}
\end{equation} 
Since $\bm{f}$ is given during training process,  $\bm{h}$ is fixed after initialization. Then this model becomes a linear combination of the basis functions $\{h_1,h_2,\dots,h_k\}$. Consequently, the nonlinear and nonconvex optimization problem of training FCNs, which is difficult to solve, is transformed into a linear least-square problem aimed at solving for $\bm{W}^{(2)}$.

\subsection{Bias Initialization Strategy}
\label{bset}

The hidden neurons in a neural network typically have the form $\sigma(\bm{Wx}+\bm{b})$. When the activation function $\sigma$ is ReLU, an important concept called the partition hyperplane, introduced in (\cite{trans}), separates the activated and unactivated regions, defined by: 
\begin{equation}
	\bm{Wx}+\bm{b}=\bm{0}.
\end{equation}

The partition hyperplane is also crucial for other activation functions, such as tanh, which is commonly used for solving PDEs. Suppose there are $M$ partition hyperplanes $\{\bm{W}_1\bm{x}+\bm{b}_1=\bm{0},\bm{W}_2\bm{x}+\bm{b}_2=\bm{0},\dots,\bm{W}_M\bm{x}+\bm{b}_M=\bm{0}\}$, Zhang et al. (\cite{trans}) further define the partition hyperplane density $D^\tau_n(\bm{x})$ as
\begin{equation}
\begin{aligned}
		D^\tau_M(\bm{x})&=\frac{1}{M}\sum_{i=1}^{M}\bm{1}_{\{d_i(\bm{x})<\tau\}}(\bm{x}),\\
		d_i(\bm{x})&=\frac{|\bm{W}_i(\bm{x})+\bm{b}_i|}{||\bm{W}_i||_2},
\end{aligned}
\end{equation}
where $\bm{1}_{\{d_i(\bm{x})<\tau\}}(\bm{x})$ is an indicator function. Zhang et al. propose that a high partition hyperplane density over the domain can improve the model’s accuracy. In this work, we adopt bias initialization strategy proposed in \cite{AdpRNN}.

Let $H = [p_1,q_1] \times \cdots \times[p_m, q_m]$ be the smallest hypercube containing the domain $D$. The RaNN model, as described by \ref{RaNN}, has $\bm{h}=\sigma(\bm{W}^{(1)}\bm{x}+\bm{b}^{(1)})$ as the output of the hidden layer. To enhance the partition hyperplane density over the domain, we initialize the bias $\bm{b}^{(1)}$ as
	\begin{equation}
		\bm{b}^{(1)}=-(\bm{W}^{(1)}\odot\bm{B})\bm{1}_{m\times1},
	\end{equation}
where $(\bm{b}_1,\cdots,\bm{b}_m)=\bm{B}\in\mathbb{R}^{k\times m}$ is generated from the uniform distribution $U(\bm{p},\bm{q})$ with $\bm{p}=(p_1,p_2,\cdots,p_m)$ and $\bm{q}=(q_1,q_2,\cdots,q_m)$.

By calculating $\bm{b}^{(1)}$ through above strategy, we enhance the model’s ability to approximate the solution space. For further details on the initialization strategy for weight parameters, based on frequency, we refer to \cite{AdpRNN}. In this paper, we focus solely on the bias initialization strategy. Developing effective initialization strategies for weight parameters specifically within the RaNN-DeepONet framework remains an intriguing and valuable topic for future research.

\section{Deep Operator Networks with Randomized Neural Networks}
\label{RaNN-DeepONet}

DeepONets, while powerful, often face optimization challenges, including high optimization errors and computational inefficiencies, stemming from the nonlinear and nonconvex nature of the optimization process. To overcome these limitations, we propose incorporating randomized neural networks (RaNNs). By taking advantage of the flexibility in the design of the branch and trunk networks, RaNNs can be seamlessly integrated into DeepONets, resulting in the RaNN-DeepONets method. In this section, we present the RaNN-DeepONets model, along with its variant, the physics-informed RaNN-DeepONets.

\subsection{Data-driven RaNN-DeepONets Model}

The RaNN-DeepONets model is data-driven, requiring no prior knowledge of PDEs, and follows the general structure of vanilla DeepONets. Consider a DeepONet model where the branch net is a RaNN, as shown in Figure \ref{fig_RNN}. The trunk net consists of a single-layer perceptron with $d$-dimensional input and $p$-dimensional output, using the $\delta(\cdot)$  as activation function. All parameters in the trunk net are randomly initialized and remain fixed throughout the training process.

\begin{figure}[h]
	\centering
	\includegraphics[width=0.8\textwidth]{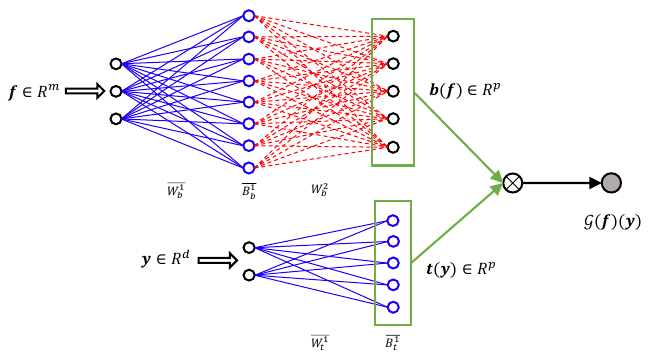}
	\caption{Architecture of RaNN-DeepONets}
	\label{fig_RaDeep}
\end{figure}

As shown in Figure \ref{fig_RaDeep}, the parameters marked with bars indicate those that are fixed during training. For the branch net, $\overline{\bm{W}_b^1}\in \mathbb{R}^{k\times m}$ and $\bm{W}_b^2\in \mathbb{R}^{p\times k}$  represent the weights of the input-to-hidden and hidden-to-output layers, respectively. The bias of the hidden layer is $\overline{\bm{B}_b^1}\in \mathbb{R}^{k}$, while the output layer bias is \textbf{0}. For the trunk net, $\overline{\bm{W}_t^1}\in \mathbb{R}^{p\times d}$ is weight and $\overline{\bm{B}_t^1}\in \mathbb{R}^{p}$ is the bias. Let $\bm{b}(\bm{f})$ and $\bm{t}(\bm{y})$ denote the outputs of branch net and trunk net, respectively. The model can then be expressed as \begin{equation*}
	\begin{aligned}
		&\mathcal{G}(\bm{f})(\bm{y})= \bm{b}(\bm{f})\cdot \bm{t}(\bm{y}),\\
		&\bm{b}(\bm{f})=\bm{W}_b^2\sigma(\overline{\bm{W}_b^1}\bm{f}+\overline{\bm{B}_b^1}),\\
		&\bm{t}(\bm{y})=\delta(\overline{\bm{W}_t^1}\bm{y}+\overline{\bm{B}_t^1}).
	\end{aligned}
\end{equation*}
The elements of $\overline{\bm{W}_b^1}$ and $\overline{\bm{W}_t^1}$ are randomly sampled from unifom distributions $U(-r_b,r_b)$ and $U(-r_t,r_t)$, respectively. $\overline{\bm{B}_t^1}$ is initialized according to the strategy discussed in Section \ref{bset}. Let $h_i$ be the output of the $i$-th neuron in the branch net's hidden layer, and $\alpha_{ij}$ be the element of $\bm{W}_b^2$ at the $i$-th row and $j$-th column. The RaNN-DeepONet model can be rewritten as:

\begin{equation}
		\mathcal{G}(\bm{f})(\bm{y})=\sum_{i=1}^{p}b_it_i(\bm{y})=\sum_{i=1}^{p}\sum_{j=1}^{k}\alpha_{ij} h_j t_i(\bm{y}).
\end{equation}

Assuming $N$ realizations of inputs are available, and for each realization, $q$ collocation points are selected to evaluate $u$, the optimization problem becomes
\begin{equation}
	\mathop{\arg\min}\limits_{\bm{\alpha}}\frac{1}{Nq}\sum_{n=1}^{N}\sum_{j=1}^{q}[\mathcal{G}(\bm{f^{(n)}})(\bm{y}^{(n)}_j) - u^{(n)}(\bm{y}^{(n)}_j)]^2,
\end{equation}
which is a linear least-square problem.

\subsection{Physics-informed RaNN-DeepONets Model}

Collecting data using traditional numerical methods is highly resource-intensive, both in terms of time and computational cost. Moreover, data-driven methods inherently face challenges such as limited generalization ability and significant memory requirements for data storage, which limit the practical applicability of RaNN-DeepONets. To overcome these challenges, we introduce the physics-informed RaNN-DeepONets model, which extends the framework of PI-DeepONets, as discussed in Section \ref{PIdeep}. The loss function of this model is divided into two components: $\mathcal{L}_{BC/IC}(\bm{\alpha})$ and $\mathcal{L}_{physics}(\bm{\alpha})$, where $\bm{\alpha}$ represents the trainable parameters of the model.

Consider the PDE given in \eqref{PIPDE}. For collocation points on the boundary, the corresponding loss $\mathcal{L}_{BC/IC}(\bm{\alpha})$ follows the form presented in \eqref{MSEloss}. For collocation points within the domain, the loss function $\mathcal{L}{physics}(\bm{\alpha})$ is computed as in \eqref{PIloss}. The model then aims to solve the following optimization problem:
\begin{equation}
	\mathop{\arg\min}\limits_{\bm{\alpha}}[\mathcal{L}_{BC/IC}(\bm{\alpha})+\mathcal{L}_{physics}(\bm{\alpha})].
\end{equation}

Furthermore, boundary conditions can be enforced as hard constraints on the physics-informed RaNN-DeepONets with only minor adjustments to the input of the trunk network. For periodic boundary conditions, the trunk net $\bm{t}(\bm{y})$ is modified to $\bm{t}(\cos(\omega \bm{y}),\sin(\omega \bm{y}))$ and its derivatives can be computed via the chain rule. For Dirichlet boundary conditions, the model takes the form:
\begin{equation}
	\mathcal{G}(\bm{f})(\bm{y})=c(\bm{y})\widetilde{\mathcal{G}}(\bm{f})(\bm{y})+g(\bm{y}),\nonumber
\end{equation}
and $\mathcal{L}_{physics}(\bm{\alpha})$ is computed as in \eqref{PIloss}.

Compared to PI-DeepONets (\cite{Pideep}), the physics-informed RaNN-DeepONets retain the advantages of the original approach while eliminating the need for automatic differentiation, as the trunk basis is fixed and its derivatives can be directly computed. As a result, physics-informed RaNN-DeepONets significantly reduce data dependency while incurring minimal additional computational cost.

\section{Numerical Examples}\label{sec:numericalexample}

In this section, we evaluate the performance of the proposed methods on three benchmark problems. Specifically, the input functions for the operators include source terms, initial conditions, and domains. The neural networks are implemented using the PyTorch library in Python, with Python version 3.11.5, PyTorch version 2.1.1, and CUDA version 12.1.

All RaNN-DeepONet models used in these experiments share the same structure. The branch network is a RaNN with a single hidden layer, where the input layer contains $m$ neurons. The widths of the hidden and output layers are $k$ and $p$, respectively. The trunk network is a single-layer perceptron using the tanh activation function. The input layer’s dimension generally corresponds to the number of independent variables in the problem, and the output layer width is $p$, matching the branch network’s output. To assess the performance of the models, we compute the relative $l^2$ error of the predictions on the test data. All RaNN-DeepONet models are trained on a single NVIDIA RTX 4060ti GPU, while the baseline models are trained on a more powerful NVIDIA V100 GPU (\cite{Pideep}).

\begin{example}[Diffusion-Reaction Dynamics]
\label{DRD}
In this example, we test the ability of RaNN-DeepONets to learn the operator mapping from the source term to the solution. We consider the following nonlinear diffusion-reaction equation with a source term $f(x)$
\begin{equation}
u_t - Du_{xx} - ku^2 = f(x), \ (x,t)\in (0,1)\times(0,1], \nonumber
\end{equation}
with zero initial and boundary conditions, where $D=0.01$ is the diffusion coefficient and $k = 0.01$ is the reaction rate. 
\end{example}

The RaNN-DeepONet is used to approximate the mapping from source term $f(x)$ to the solution $u(x,t)$, i.e.,
\begin{equation}
\mathcal{G}: f(x)\rightarrow u(x,t).\nonumber
\end{equation}

The dataset used for training is taken from \cite{Pideep}, containing $N=10,000$ random realizations of $f(x)$, generated by sampling a Gaussian random field (GRF) with length scale $l=0.2$, along with their corresponding numerical solutions. Each solution has a resolution of $100\times100$, and all $u$-values required for training are obtained through a linear interpolation on this mesh. The test data contains another $1000$ realizations evaluated on a $100\times100$ uniform grid.

To discretize the inputs, we set $m=100$, where $\{x_1,x_2,\cdots,x_m\}$ is uniformly distributed in the interval \( [0, 1] \) and $q=100$ points are randomly selected from the \( [0, 1] \times [0, 1] \) domain. The RaNN-DeepONet model is built with $k=120$, $p=100$, $r_b=0.003$ and $r_t=8$. In each training process, $40,000$ data points are randomly selected. We compare the results with those obtained by the vanilla DeepONet and the physics-informed DeepONet from Wang et al. (\cite{Pideep}), where both their branch and trunk networks consist of 5-layer fully connected neural networks with 50 neurons per layer.

As discussed in Section \ref{HBC}, we enforce the zero Dirichlet BC on the RaNN-DeepONet by choosing the surrogate solution as
\begin{equation}
	\mathcal{G}(\bm{f})(x,t)=c(x,t)\widetilde{\mathcal{G}}(\bm{f})(x,t),\nonumber
\end{equation}
where $c(x,t)=tx(1-x)$ and $\widetilde{\mathcal{G}}(\bm{f})(x,t)$ is the RaNN-DeepONet solution.

\begin{table}[h]
	\centering
\begin{tabular}{|l|l|l|l|}
\hline
\multirow{1}{*}{Model}    & Parameters    & \multicolumn{1}{c|}{Time (s)} & \multicolumn{1}{c|}{Average Relative $l^2$ Errors}                     \\ 
\hline
\multirow{1}{*}{Vanilla DeepONet}    & $25100$    & \multicolumn{1}{l|}{$4068.00$}  & \multicolumn{1}{l|}{1.92E-02}                           \\ 
\hline
\multirow{1}{*}{Physics-Informed DeepONet}    & $25100$    & \multicolumn{1}{l|}{$8172.00$}  & \multicolumn{1}{l|}{4.50E-03}                           \\ 
\hline
\multirow{1}{*}{RaNN-DeepONet}    & $12000$    & \multicolumn{1}{l|}{$173.21$}  & \multicolumn{1}{l|}{5.40E-03}                           \\ 
\hline
\multirow{1}{*}{RaNN-DeepONet (HCB)}    & $12000$    & \multicolumn{1}{l|}{$90.74$}  & \multicolumn{1}{l|}{2.80E-03}                           \\ 

\hline
\end{tabular}
\caption{Parameters, training time and errors of different models in Example \ref{DRD}}
\label{tableDRD}
\end{table}

\begin{figure}[h]
	\centering
	\subfloat[RaNN-DeepONet]{
		\includegraphics[scale=0.3]{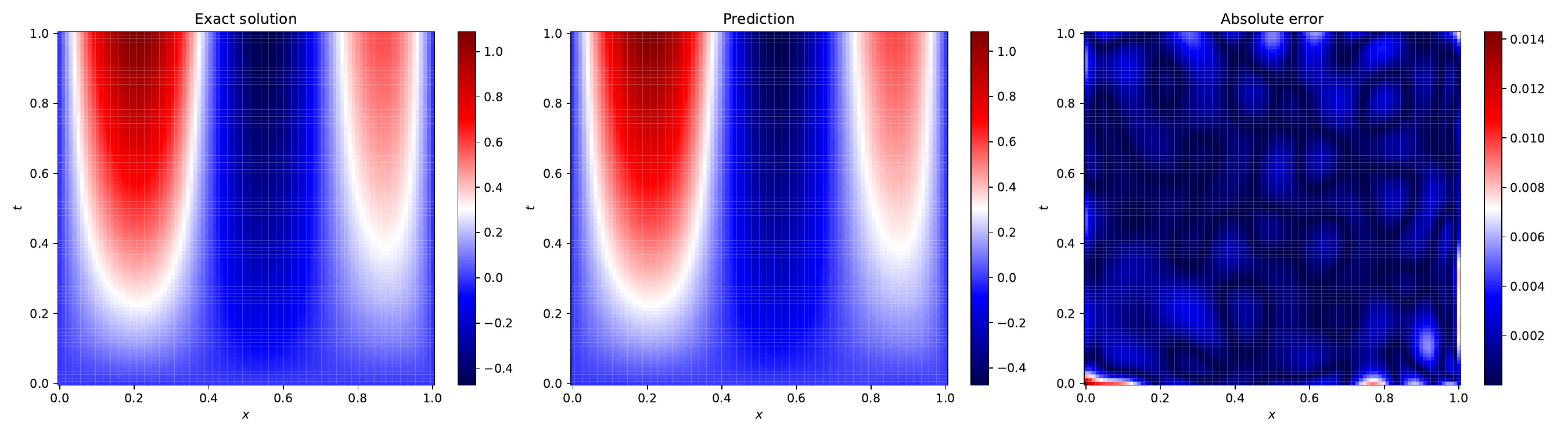}}\\
	\subfloat[RaNN-DeepONet with hard-constraint]{
		\includegraphics[scale=0.3]{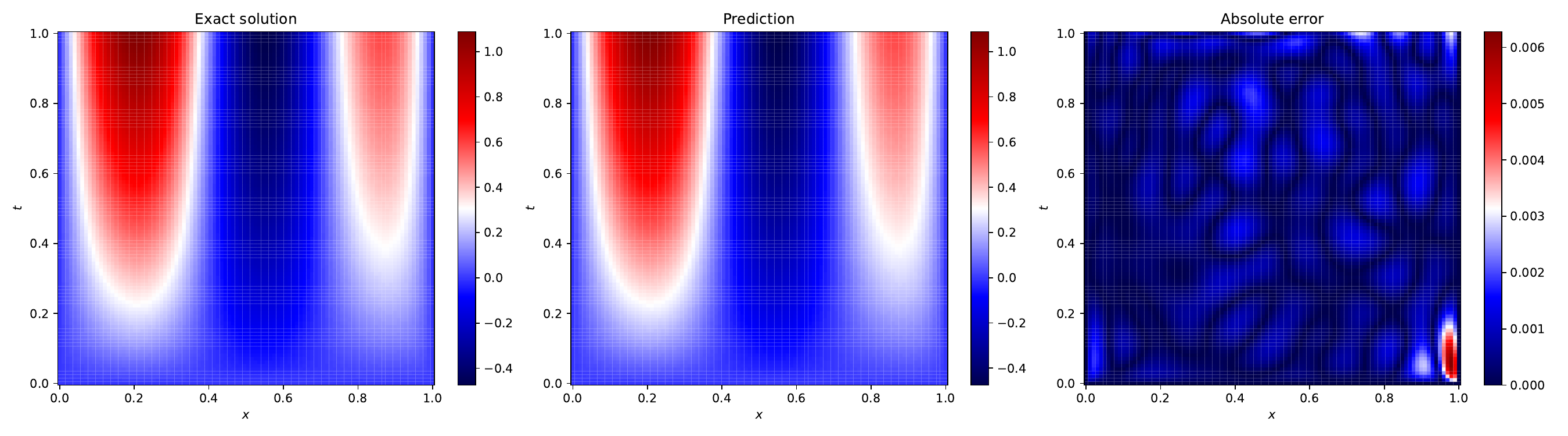}}
	\caption{Example of exact solutions, predictions, and absolute errors in Example \ref{DRD}}
	\label{fig_DRD}
\end{figure}

Table \ref{tableDRD} compares the performance of Vanilla DeepONet, Physics-Informed DeepONet, RaNN-DeepONet, and RaNN-DeepONet with a hard constraint. It can be observed that, compared to the Vanilla DeepONet, the RaNN-DeepONet method achieves significantly higher accuracy while requiring much less training time. Although the Physics-Informed DeepONet achieves better accuracy, its training speed is slower due to the auto-grad implementation for computing second-order derivatives.

To further improve the model’s accuracy, we introduce the hard constraint for Dirichlet boundary conditions, as discussed earlier. By introducing a simple function \( c(x,t) \), the accuracy of RaNN-DeepONet is improved, outperforming all other models. The incorporation of hard-constraint boundary conditions enables the neural network to automatically satisfy the boundary conditions of the problem, reducing boundary errors and enhancing the approximation capability of the trunk network. 

We use the trained RaNN-DeepONet and the RaNN-DeepONet model with hard constraints to solve a specific diffusion-reaction equation. Figure \ref{fig_DRD} displays the predicted solutions and absolute errors for both methods. Figure \ref{fig_DRD} (a) shows the results of the RaNN-DeepONet, with a relative \( l_2 \) error of $0.0042$. Figure \ref{fig_DRD} (b) shows the results of the RaNN-DeepONet with hard constraints, yielding a relative \( l_2 \) error of $0.0019$. The comparison between the two figures reveals that enforcing Dirichlet boundary conditions significantly reduces the solution error, particularly at the domain boundaries.

\begin{example}[Burgers' Equation]
\label{BE}

To demonstrate the ability of RaNN-DeepONets in learning the operator mapping from initial condition to the solution across the entire domain, we apply this method to solve Burgers’ equation. Consider the one-dimensional Burgers’ equation:
\begin{equation*}
\begin{aligned}
u_t+uu_x-\nu u_{xx} &= 0, (x,t)\in (0,1)\times (0,1],\\ \nonumber
u(x,0) &= u_0(x), \nonumber
\end{aligned}
\end{equation*}
with periodic boundary conditions,
\begin{equation*}
	\begin{aligned}
		u(0,t)&=u(1,t),\\ \nonumber
		u_x(0,t)&=u_x(1,t),\nonumber
	\end{aligned}
\end{equation*}
where $t\in (0,1)$, and the viscosity is set to $\nu=0.01$.
\end{example}

The goal is to use RaNN-DeepONets to approximate the operator that maps the initial condition to the solution $u(x,t)$, i.e.,
\begin{equation}
	\mathcal{G}: u_0(x)\rightarrow u(x,t).\nonumber
\end{equation}

\begin{figure}[h]
	\centering
	\includegraphics[width=0.6\textwidth]{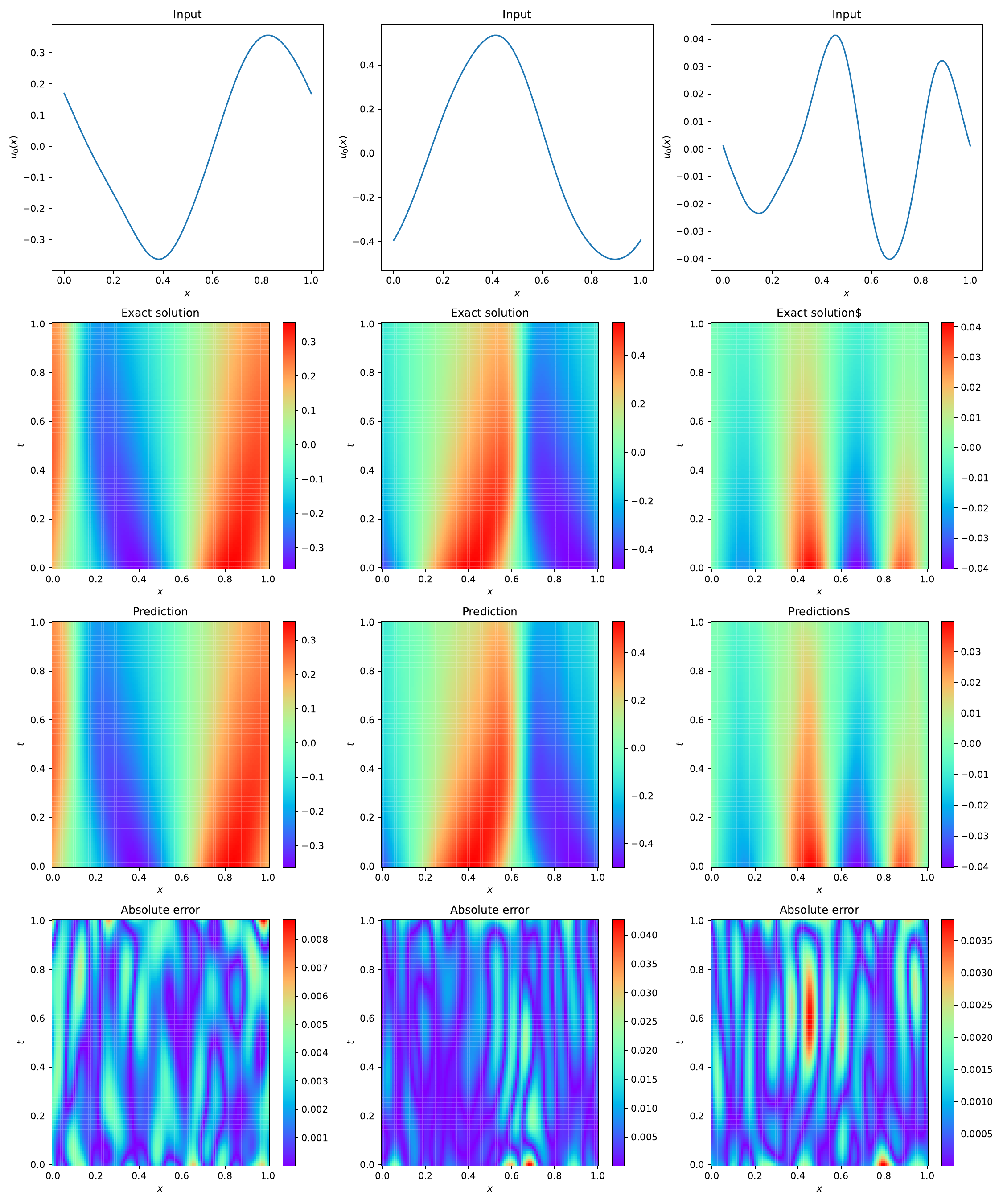}
	\caption{Examples of inputs, exact solutions, predictions, and absolute errors in Example \ref{BE} (including the worst case)}
	\label{fig_BE}
\end{figure}

\begin{figure}[h]
	\centering
	\includegraphics[width=0.6\textwidth]{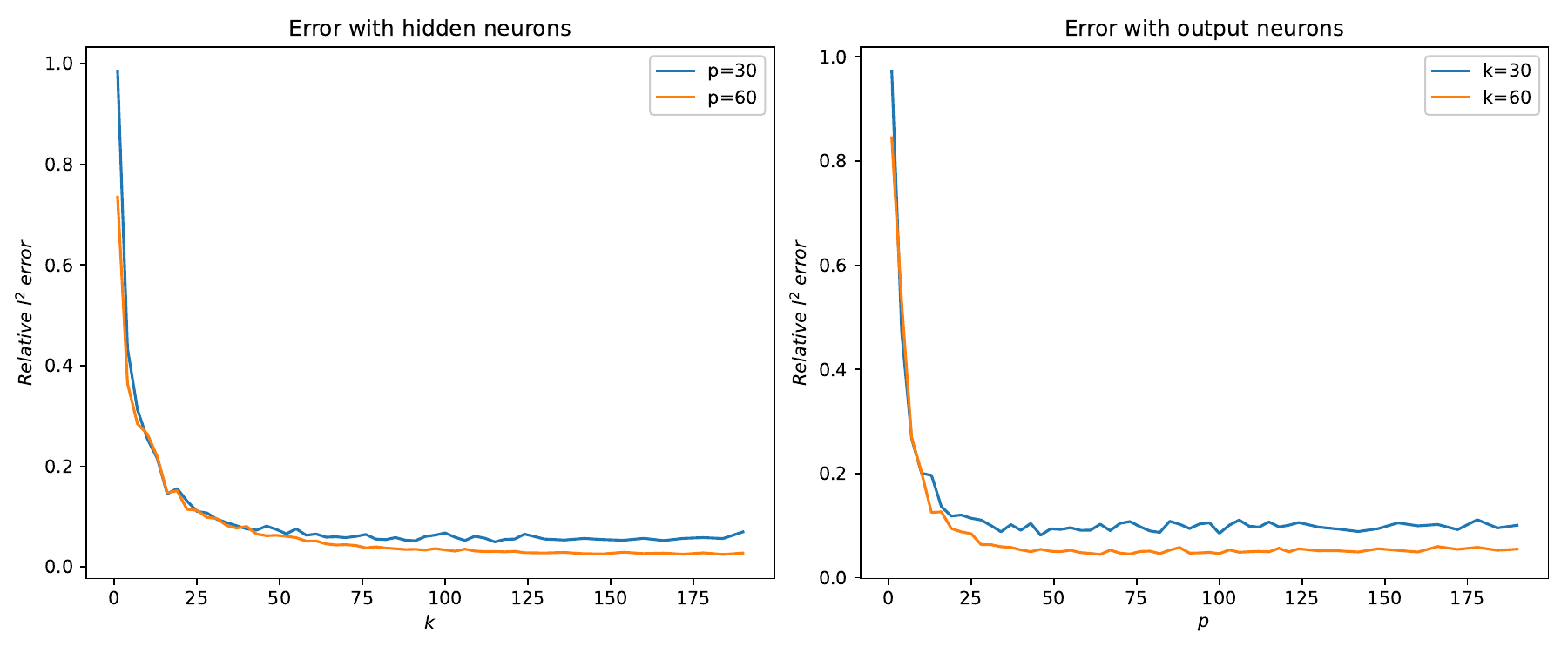}
	\caption{The effects of hidden and output layer neurons on model's accuracy}
	\label{fig_BE_neurons}
\end{figure}

We use the code from Lu et al. (\cite{PODdeep}) to generate training data. This generates $N=1000$ random realizations of $u_0(x)$ and their corresponding numerical solutions. The initial condition $u_0(x)$ is generated from a GRF with a Riesz kernel that satisfies the periodic BCs, denoted as $\mu \sim \mathcal{R}(0,625(-\Delta+25I)^{-4})$. The test data cconsists of another $100$ realizations evaluated on a $101\times101$ uniform grid, generated using the same method as the training data.

Following the experimental setup of Wang et al. (\cite{Pideep}), we set $m=101$, discretizing each $u_0(x)$ into $101$ uniformly distributed points in $[0,1]$. Then, we select $q = 2,701$ collocation points. Among these, $101$ points are sampled on the initial boundary at $t = 0$, $100$ points are distributed on the left and right boundaries of $x$-direction ($x = 0$ and $x = 1$, with $50$ points allocated to each boundary), and $2,500$ points are randomly selected from the remaining interior domain. 

As the baseline for this experiment, Wang et al. use two separate 7-layer FCNs to represent the branch net and the trunk net, respectively. Each network uses tanh as activation function and has 100 neurons per hidden layer. For the RaNN-DeepONet, we set $k=200$, $p=120$, $r_b=0.5$ and $r_t=2$, and randomly select $86,432$ data points in training process, with the proportion consistent with the dataset.

To handle periodic boundary conditions, we apply the feature expansion method, as mentioned in Section \ref{HBC}. By applying the Fourier basis \{$\cos (2\pi x)$,$\sin (2\pi x)$\} to the spatial variable of the collocation points, the input of the trunk net is expanded to \{$\cos (2\pi x)$,$\sin (2\pi x), t$\}, ensuring that the model’s output satisfies periodic boundary conditions. The training process takes 289.00s, with an average relative $l^2$ error of 0.0157. In comparison, the baseline model, PI-DeepONet, achieves an accuracy of 0.0138 but requires 27,396.00s, demonstrating that the proposed model provides comparable accuracy with significantly reduced training time.

Figure \ref{fig_BE} illustrates the solution accuracy of the trained RaNN-DeepONet model for Burgers’ equation under three distinct initial conditions. The four rows display the initial condition as input to the branch net, the exact solution, the model prediction, and the absolute error, respectively. The relative $l^2$ errors for these cases on the $101\times101$ uniform grid are 0.0114, 0.0270, and 0.0778. The third input corresponds to the test sample with the worst prediction accuracy on the trained model. Notably, even in this worst-case scenario, the model’s prediction shows excellent agreement with the exact solution.

As previously mentioned, the number of neurons in the hidden and output layers of the branch net significantly influences the experimental results. To investigate this, we fix the hidden layer width $k$ and the output layer width $p$, respectively, and vary the other parameters to observe any changes in model accuracy.

Figure \ref{fig_BE_neurons} (a) shows the test error variation with hidden layer neurons when fixing the output layer width to $p=30$ and $p=60$, respectively. A significant discrepancy in accuracy emerges when $k>50$ , with a minimum relative $l^2$ error of 0.0494 for $p=30$ and 0.0245 for  $p=60$. Similarly, Figure \ref{fig_BE_neurons} (b) demonstrates the test error variation with changing output layer neurons under fixed hidden layer widths $k=30$ and $k=60$. Distinct performance differences are observed, yielding minimum errors of 0.0813 and 0.0444, respectively.

For the output layer width of the branch network, $p$ should be aligned with the number of neurons in the trunk network. A larger $p$ implies more trunk basis functions, providing a greater chance for the corresponding space to encompass the target operator. One might assume that model accuracy could be improved by excessively increasing $p$. However, Figure \ref{fig_BE_neurons} (b) demonstrates that when $k$ is small, simply increasing $p$ does not improve model accuracy. Therefore, we emphasize that the scales of $k$ and $p$ should be comparable during the training process.

\begin{example}[Darcy problem]
	\label{DF}
This example demonstrates the capability of the proposed RaNN-DeepONet model and its physics-informed variant in solving problems involving diverse domains as input. Consider the two-dimensional Darcy flow in various geometries filled with porous media
	\begin{equation}
		-\nabla \cdot (K\nabla u(x,y))=f,\quad (x,y)\in \Omega,  \nonumber
	\end{equation}
subject to homogenous Dirichlet boundary condition, where $K = 1$ represents the permeability field, $u$ is the pressure, and $f = 1$ is the source term.\end{example}

Our goal is to use the proposed model to learn the operator mapping $\Omega$ to the solution $u(x,y)$, i.e.,
 \begin{equation}
 	\mathcal{G}: \Omega \rightarrow u(x,y).\nonumber
 \end{equation}

\noindent \textbf{Data  generation.} We employ MATLAB’s Partial Differential Equation Toolbox to solve Darcy problems across diverse domains. To evaluate the RaNN-DeepONet’s ability to handle different geometries, we generated $N=2700$ realizations with three domain types: ellipses, rectangles, and isosceles triangles (900 instances per shape), denoted as $\{\Omega^{(1)}, \Omega^{(2)}, \dots, \Omega^{(n)}\}$. These domains are subsets of a maximal rectangular domain $\mathcal{D} = [0, 2] \times [0, 2]$. For the boundary of each domain $\Omega^{(n)}$, we discretize it into 100 uniformly distributed points as $d\Omega^{(n)}$ and sample $q = 1000$ collocation points on $\mathcal{D}$. The domain’s parametric information $\Omega^{(n)}_{par}$, which varies depending on the domain shape, is included but not used as input to the branch net. Thus, the training data is denoted as $\{\{d\Omega^{(n)}_i\}^{m}_{i=1},\Omega^{(n)}_{par},\{y^{(n)}_j,u^{(n)}_j\}^q_{j=1}\}^N_{n=1}$. During training, we flatten $d\Omega^{(n)}$ into a vector as input to the branch net. The test data consists of another 300 realizations (100 per shape), evaluated on a $201 \times 201$ uniform grid of $\mathcal{D}$.

Notably, when $\Omega^{(n)}$ is small, the number of collocation points $q$ must be sufficiently large to ensure adequate coverage within the domain, which increases memory demands. To mitigate this, we prioritize larger domain sizes during the domain generation process, thus reducing memory pressure. The random parameters governing the domain generation are as follows:
\begin{itemize}
\item \textit{Ellipse}: The center coordinates of the ellipse, $x_c$ and $y_c$, are sampled from a uniform distribution within $[0.8, 1.2] \times [0.8, 1.2]$. The lengths of the major and minor axes, $l_{maj}$ and $l_{min}$, are drawn from $U(0.3, 0.8)$, and the tilt angle $\theta$ is sampled from $U(0, 2\pi)$.
\item \textit{Rectangle}: Four real numbers are sampled from $U(0, 1)$ and sorted in ascending order, denoted by $\{x_0, y_0, x_1, y_1\}$. To ensure the rectangle domain is large enough, we use $\{x_{min} = x_0, y_{min} = y_0, x_{max} = x_1 + 1, y_{max} = y_1 + 1\}$ to generate the rectangle.
\item \textit{Isosceles Triangle}: The coordinates of the vertex are $(x_v, y_v)$, where $x_v = 1$ and $y_v \sim U(1.5, 2.0)$. The height $h$, vertical to the $x$-direction, is sampled from $U(0.9, 1.5)$. The length of the base edge, $b$, is drawn from $U(1.2, 2.0)$.
\end{itemize}

\noindent \textbf{Hard-constraint.} To handle different shapes during the training process, surrogate solutions are required for each $\Omega_{par}$. The primary challenge is determining whether a collocation point lies within the target domain. We design three functions, $c(x, y)$, to address this for ellipses, rectangles, and isosceles triangles, respectively.

For the \textit{ellipse}, the parametric form of $\Omega_{par}$ is $\{type=0, (x_c, y_c), l_{maj}, l_{min},\theta\}$, where $type = 0$ represents an ellipse. We define $c(x, y)$ as:
\begin{equation*}
	\begin{aligned}
		c(x,y)&=\left\{  \begin{aligned}
			&j(x,y), \quad j(x,y)>0\\
			&0, \qquad\quad\; j(x,y)\leq0
		\end{aligned}\right.
	\end{aligned}
\end{equation*}
where
\begin{equation*}
		j(x,y)=1-\left(\frac{(x-x_c)\cos \theta+(y-y_c)\sin \theta}{x_l}\right)^2-\left(\frac{(y-y_c)\cos \theta-(x-x_c)\sin \theta}{y_l}\right)^2.
\end{equation*}
Here, a value of $j(x, y) > 0$ indicates that the point is inside the ellipse, while $j(x, y) < 0$ means the point lies outside the ellipse, and we set it to $0$ to satisfy the Dirichlet boundary conditions (BCs).

For the \textit{rectangle}, the parametric form of $\Omega_{par}$ is $(type = 1, x_{min}, y_{min}, x_{max}, y_{max})$, and we can easily determine the position relationship between a point and the rectangular domain. We define $c(x, y)$ as:
\begin{equation*}
		c(x,y)=\left\{  \begin{aligned}
			&(x-x_{min})(x-x_{max})(y-y_{min})(y-y_{max}),\ (x,y) \in \Omega\\
			&0, \qquad\qquad\qquad\qquad\qquad\qquad\qquad\qquad\qquad\; {\rm otherwise}
		\end{aligned}\right.
\end{equation*}

For the \textit{isosceles triangle}, the parametric form of $\Omega_{par}$ is $\{type = 2, (x_v, y_v), (x_{lb}, y_{lb}), (x_{rb}, y_{rb})\}$, where $(x_{lb}, y_{lb})$ and $(x_{rb}, y_{rb})$ are the coordinates of the left and right base angles of the isosceles triangle. We can compute the parametric equations for the three sides of the triangle: $l_1(x, y)$, $l_2(x, y)$, and the bottom side $l_3(x, y)$. If the value is greater than 0, the point lies above the line; if it is less than 0, the point lies below the line. The domain is parameterized as:
\begin{equation*}
	c(x,y)=\left\{  \begin{aligned}
		&l_1(x,y)l_2(x,y)l_3(x),\ (x,y) \in \Omega\\
		&0, \qquad\qquad\qquad\qquad {\rm otherwise}
	\end{aligned}\right.
\end{equation*}

These three forms of $c(x, y)$ ensure the hard-constraint Dirichlet BCs are satisfied, allowing for the use of the corresponding surrogate solutions during experiments.

\begin{figure}[h]
	\centering
	\includegraphics[width=0.7\textwidth]{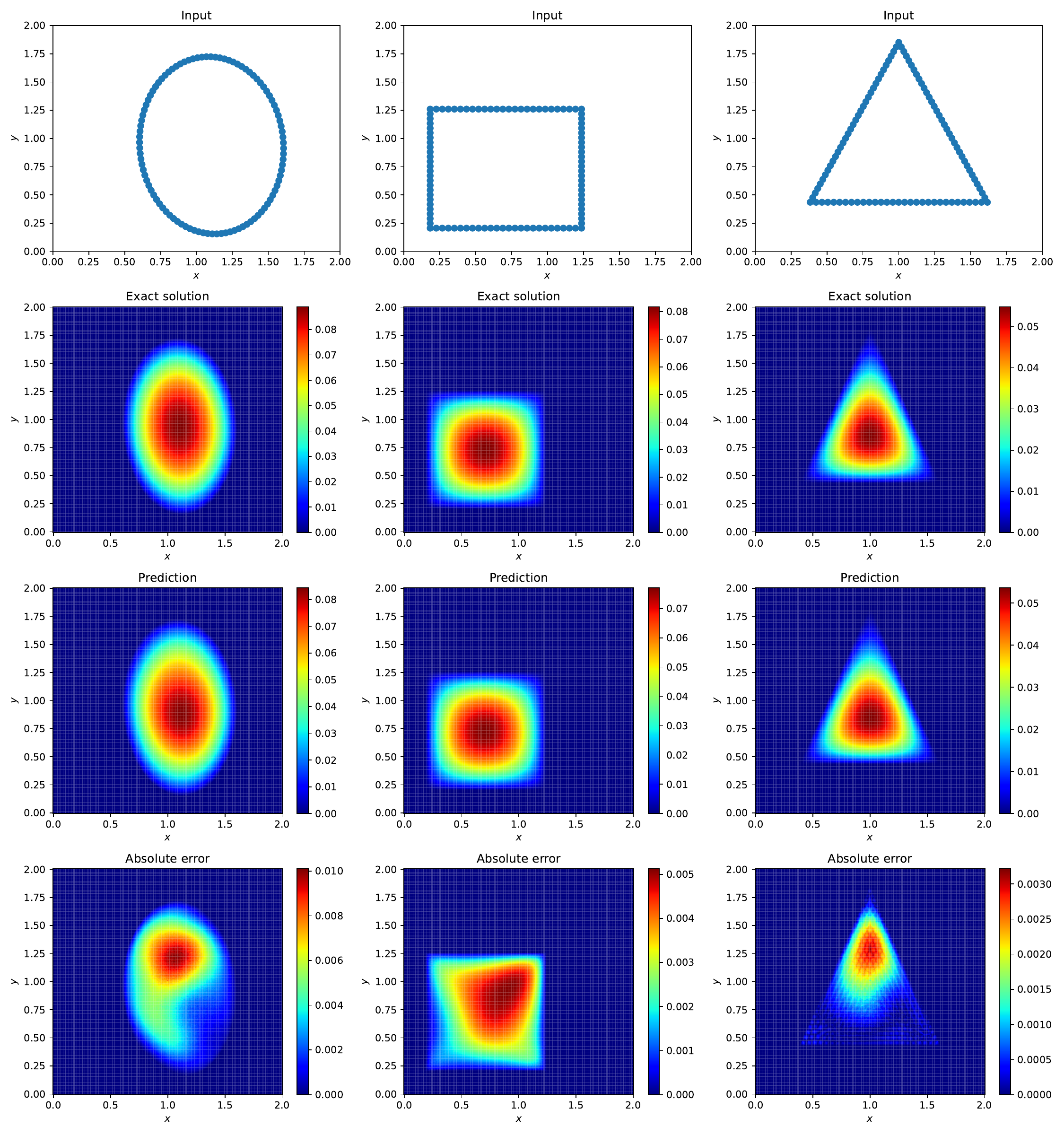}
	\caption{Worst-case performance of each domain type with RaNN-DeepONet in Example \ref{DF}}
	\label{fig_DF}
\end{figure}

\noindent \textbf{RaNN-DeepONets.} We construct the RaNN-DeepONet with hyperparameters set as $k=200$, $p=60$, $r_b=0.05$ and $r_t=1$. During the training process, $150,000$ data points are randomly sampled, resulting in a training duration of $243.55s$ and achieving an average relative $l^2$ error of 0.0188. Figure \ref{fig_DF} presents the worst-case predictions made by the trained RaNN-DeepONet model for three distinct domain shapes, evaluated on a $201\times201$ uniform grid. The four rows of the figure correspond, respectively, to the discretized domain used as input to the branch net, the exact solution, the model’s predicted solution, and the absolute error. The worst-case errors for the three shapes are
0.0858, 0.0666, and 0.0382, respectively. Even in these worst-case scenarios, the proposed model demonstrates excellent agreement with the true solutions. This result underscores the capability of RaNN-DeepONet to accurately recognize and handle diverse domains with significant geometric variations, highlighting its strong potential in approximating region-dependent PDE operators.

\noindent \textbf{Physics-informed RaNN-DeepONets.} Compared to the RaNN-DeepONet, the physics-informed variant removes the need for explicit solution values $u$ in the training data. Specifically, for this Darcy problem with a constant source term $f=1$ and homogeneous Dirichlet boundary conditions, the physics-informed RaNN-DeepONet training dataset consists solely of the domain boundary points and their parameterizations: $\{\{d\Omega^{(n)}_i\}^{m}_{i=1},\Omega^{(n)}_{par},\{y^{(n)}_j\}^q_{j=1}\}^N_{n=1}$. The hyperparameters for the physics-informed RaNN-DeepONet are selected as $k=150$, $p=60$, $r_b=0.05$ and $r_t=2$, with $120,000$ data points sampled for training. This training process takes only 99.00s and yields an average relative $l^2$ error of 0.0203.

\begin{figure}[h]
	\centering
	\includegraphics[width=0.7\textwidth]{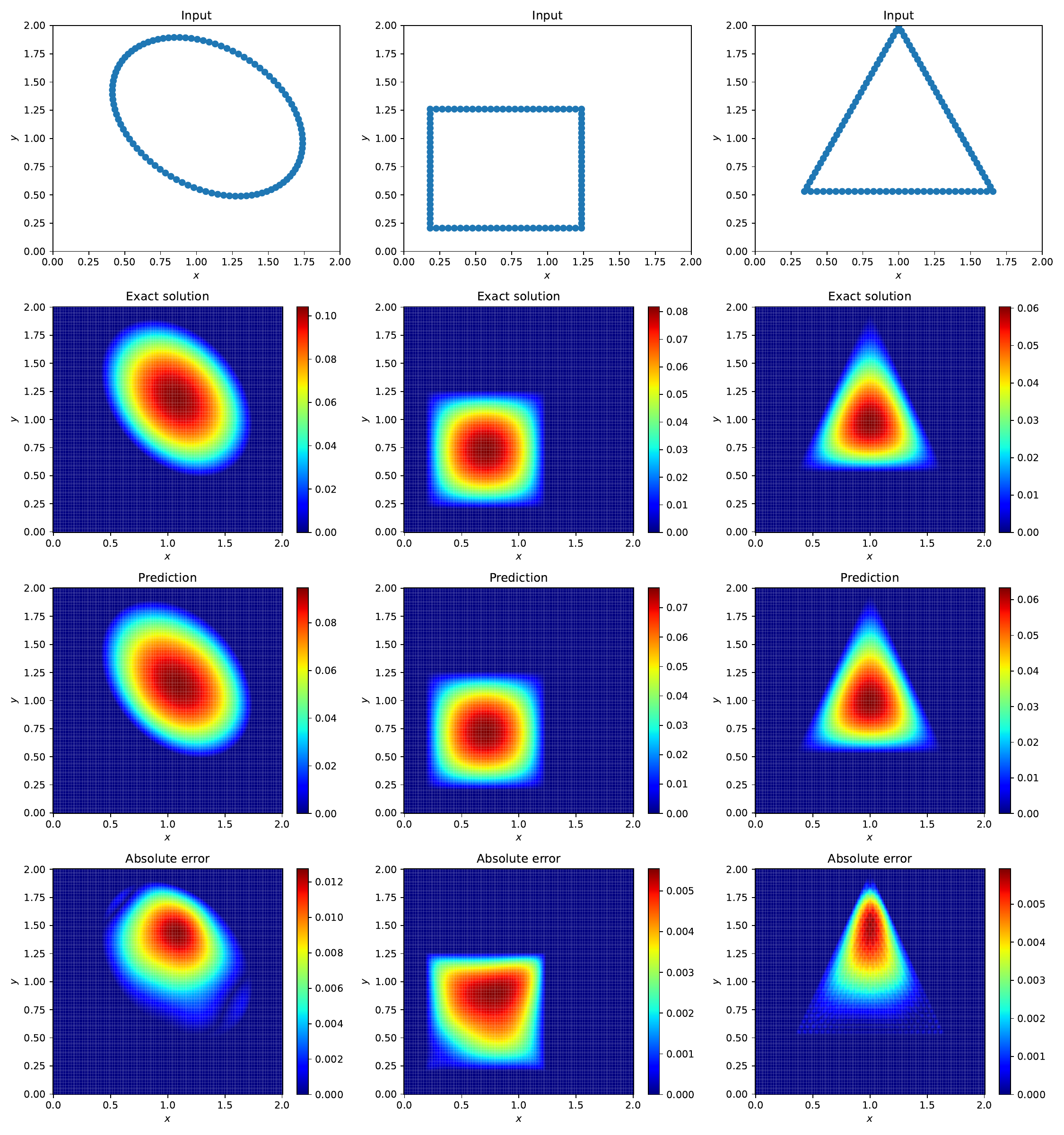}
	\caption{Worst-case performance of each domain type with physics-informed RaNN-DeepONet in Example \ref{DF}
	}
	\label{fig_PIDF}
\end{figure}

Figure \ref{fig_PIDF} illustrates the worst-case predictions of the trained physics-informed RaNN-DeepONet across the three different domain shapes. The resulting errors for these worst-case scenarios are 0.0953, 0.0708, 0.0773, respectively. As clearly demonstrated, the physics-informed RaNN-DeepONet matches the accuracy of the RaNN-DeepONets while requiring fewer parameters. Additionally, because the trunk network utilizes differentiable fixed basis functions (such as tanh) derivatives required in the training process can be explicitly computed without resorting to automatic differentiation, significantly reducing computational time compared to conventional physics-informed DeepONets (\cite{Pideep}).

\section{Summary}\label{sec:summary}

The RaNN-DeepONet proposed in this paper offers an efficient and accurate approach for learning nonlinear PDE operators, substantially reducing the computational resources required during training. Built upon the foundational DeepONet framework, our method integrates the concept of RaNNs into its architecture. Specifically, we employ a RaNN structure for the branch network, while replacing the original DeepONet trunk network with a fixed set of basis functions. Consequently, the complex nonlinear and nonconvex optimization encountered in training a conventional DeepONet is converted into a linear optimization problem, efficiently solvable via the least-squares method.

We validate the performance of our RaNN-DeepONet framework through three representative PDE examples: diffusion-reaction dynamics, Burgers’ equation, and Darcy flow problems. These experiments illustrate the model’s ability to effectively approximate diverse operators, mapping source terms, initial conditions, and geometric domains to their respective PDE solutions. The results highlight two notable advantages: (i) the RaNN-DeepONet achieves an accuracy comparable to the vanilla DeepONet, and even outperforms it significantly in the diffusion-reaction dynamics example; (ii) due to the adoption of the least-squares method for training, RaNN-DeepONet requires substantially fewer computational resources and greatly reduces the training time compared to the traditional DeepONet.

Despite the promising results demonstrated by RaNN-DeepONet in operator learning tasks, several aspects of the method merit further exploration. Fundamentally, RaNN-DeepONet remains a data-driven approach, heavily reliant on extensive numerical solutions for training. To address this limitation, we integrate physics-informed techniques into the RaNN-DeepONet framework and apply this enhanced model to the Darcy problem. Thanks to the fixed basis functions employed in its trunk network, RaNN-DeepONet allows direct computation of derivatives, thereby eliminating the computational overhead typically associated with automatic differentiation. Moreover, the physics-informed RaNN-DeepONet significantly reduces data dependency, a benefit clearly demonstrated by our successful numerical experiments involving the Darcy flow problem.
However, directly extending this benefit to strongly nonlinear PDEs such as the diffusion-reaction and Burgers’ equations poses additional challenges. Possible approaches to resolve this issue include linearizing these nonlinear equations using techniques like Newton’s method or Picard iteration.

Additionally, enhancing the capability of RaNN-DeepONet to address increasingly complex problems represents an important research direction. A particularly promising approach involves exploring the integration of advanced neural network structures—such as convolutional neural networks or transformer-based neural networks—into the RaNN-DeepONet framework. These directions will form the foundation for our future studies, guiding further development, refinement, and expansion of the RaNN-DeepONet framework.

\end{document}